\documentclass[11pt,a4paper]{article}
\usepackage[hyperref]{eacl2021}
\usepackage{times}
\usepackage{latexsym}

\usepackage{microtype}

\aclfinalcopy % Uncomment this line for the final submission
\def\aclpaperid{32} %  Enter the acl Paper ID here

\usepackage{graphicx}

\usepackage{array}  % for fixed table width
\newcolumntype{L}[1]{>{\raggedright\let\newline\\\arraybackslash\hspace{0pt}}p{#1}}
\newcolumntype{R}[1]{>{\raggedleft\let\newline\\\arraybackslash\hspace{0pt}}p{#1}}

\newcommand*{\affaddr}[1]{#1} % No op here. Customize it for different styles.
\newcommand*{\affmark}[1][*]{\textsuperscript{#1}}
\newcommand*{\email}[1]{\texttt{#1}}

\title{
Towards a parallel corpus of Portuguese and the Bantu language Emakhuwa of Mozambique
}

\author{%
Felermino D. M. A. Ali\affmark[1], Andrew Caines\affmark[2], Jaimito L. A. Malavi\affmark[1]\\
\affaddr{\affmark[1]Department of Computer Engineering, Lurio University, Mozambique}\\
\affaddr{\affmark[2] Computer Laboratory, University of Cambridge, United Kingdom}\\
\email{\{felermino.ali,jmalave\}@unilurio.ac.mz},\\\email{andrew.caines@cl.cam.ac.uk}\\
}
\date{}

\begin{document}
\maketitle

\begin{abstract}
Major advancement in the performance of machine translation models has been made possible in part thanks to the availability of large-scale parallel corpora. 
%Machine translation in low-resource languages still is a challenge and more research needs to be done for the development of methods as well as the creation of more resources. 
But for most languages in the world, the existence of such corpora is rare.
Emakhuwa, a language spoken in Mozambique, is like most African languages low-resource in NLP terms. It lacks both computational and linguistic resources and, to the best of our knowledge, few parallel corpora including Emakhuwa already exist. In this paper we describe the creation of the Emakhuwa-Portuguese parallel corpus, which is a collection of texts from the Jehovah's Witness website and a variety of other sources including the African Story Book website, the Universal Declaration of Human Rights and Mozambican legal documents. The dataset contains 47,415 sentence pairs, amounting to 699,976 word tokens of Emakhuwa and 877,595 word tokens in Portuguese. After normalization processes which remain to be completed, the corpus will be made freely available for research use.
\end{abstract}

\section{Introduction}

Machine translation (MT) is the process by which computational models are trained to transform a source language text into a target language text, and is a technology which in recent years has seen great improvements in performance thanks to the development of MT with neural networks (NMT) \cite{kalchbrenner-blunsom-2013,sutskever-2014,bahdanau-2015}. NMT typically depends on supervised machine learning from large parallel corpora of texts in the source and target language pair. The creation of such corpora usually depends on extrinsic motivation for the existence of abundant parallel text: for instance, government activity in multilingual settings such as the European and Canadian Parliaments. In the context of NMT, this fact has led to a dichotomous situation of high and low-resource language pairings. For the academic community, English-German or English-French are prototypical high-resource translation pairs whereas most other language pairs are low-resource in comparison.

We present a corpus for a low-resource language pair: Portuguese and Emakhuwa (alternatively, `Makhuwa' or `Makua'), the official and the most widely spoken languages of Mozambique, respectively.
Emakhuwa is spoken in all the provinces of northern Mozambique, namely Niassa, Cabo Delgado and Nampula and also in Zambezia in central northern Mozambique. It is estimated that approximately 25\% of the country's population of 30 million people make use of the language on a daily basis as an alternative to Portuguese \cite{paula-duarte-2016}. Emakhuwa is also spoken in some of the neighbouring countries to the north of Mozambique \textendash~namely, Tanzania and Malawi \cite{ngunga-2012} but speaker populations in these countries are relatively small.

\begin{figure*}
    \centering
    \includegraphics[width=.8\textwidth]{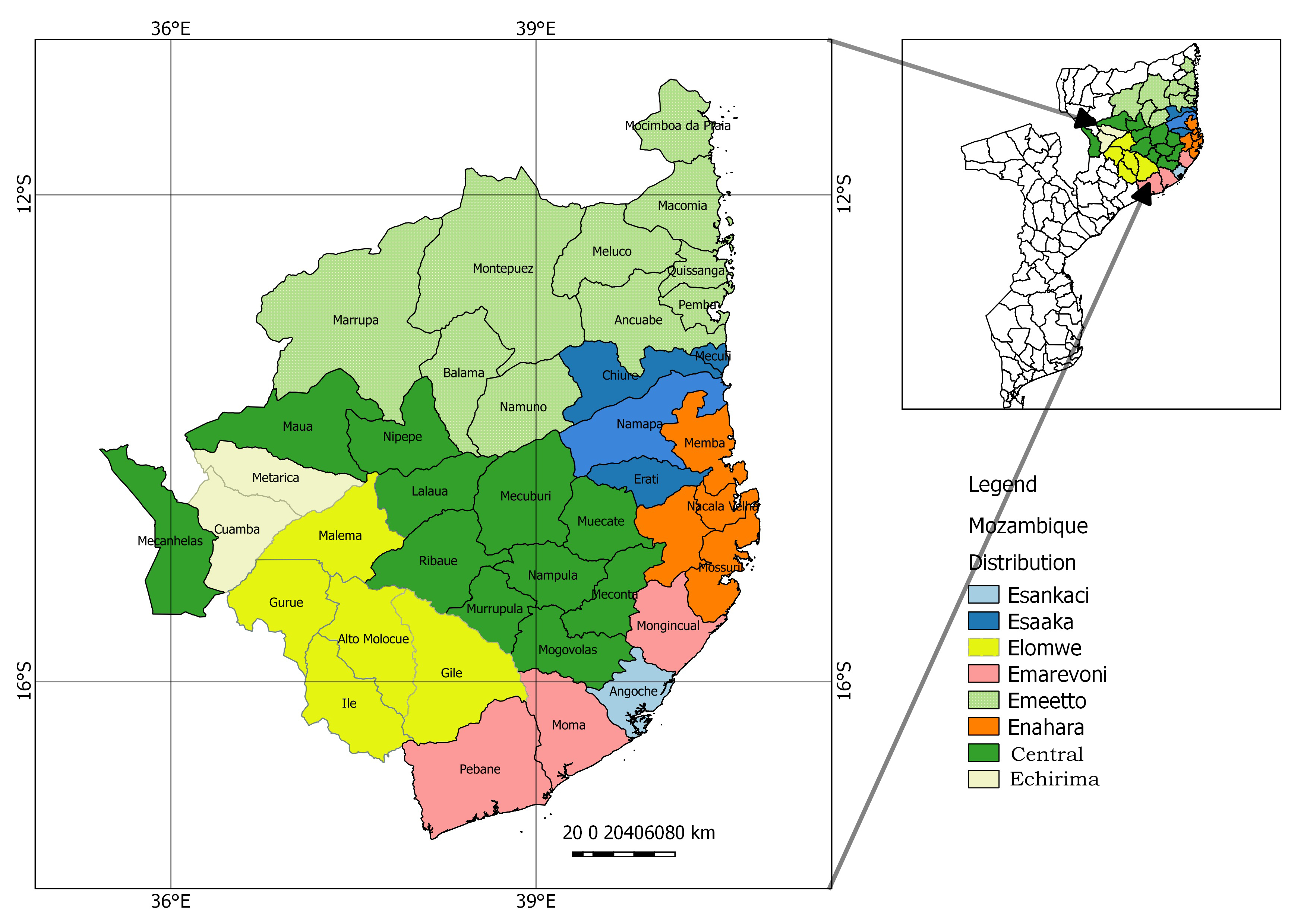}
    \caption{Geographic distribution of Emakhuwa variants in Mozambique \cite{ngunga-faquir-2012}.}
    \label{fig:variants}
\end{figure*}

There are 8 variants of Emakhuwa \cite{ngunga-faquir-2012}: Elomwe (ISO-639 code: ngl), Esankaci, Esaaka (ISO-639 code: xsq), Echirima (ISO-639 code: mhm), Emarevoni (ISO-639 code: xmc), Emeeto (ISO-639 code: mgh), Enahara and Central (ISO-639 code: vmw).
The variants are distributed across different districts of northern and central Mozambique (see Figure~\ref{fig:variants}) and differ slightly in terms of accents and lexicon.

Like many languages spoken on the African continent, Emakhuwa has limited resources for computational linguistics compared with English, French, Portuguese, and so on. Thus we are collecting a dataset for MT in which Portuguese texts are paired with Emakhuwa translations. The 47,415 sentence pairs we have collected contain 699,976 word tokens of Emakhuwa and 877,595 word tokens in Portuguese. The corpus will be made available for research use when final normalization and data collection has been completed\footnote{\url{placeholder.com}}. In addition we will seek out new data sources and continue to expand the corpus.

\section{Related Work}
\label{litrev}

In general, African languages have been relatively little explored by the NLP research community. However, lately this scenario seems to be improving as more data are being made available for research use, including translation work. OPUS \cite{Tiedemann-2012}, the 1000Langs corpus crawled from Bible corpora \cite{asgari-schutze-2017-past}, and the JW300 project \cite{agic-vulic-2019-jw300} are examples of this trend, as they made available parallel corpora of over 300 languages including Emakhuwa-Portuguese. 
%Nevertheless, their data were extracted for JW's website which beside been religious biased it may not generalize well, as the word conflicts may happen when trying to translate sentences from other sources. 
The Crúbadán Project\footnote{\url{http://crubadan.org/languages/vmw}} also  used Jehovah's Witness resources to create a corpus of Emakhuwa. However, the corpus is monolingual and relatively small with around 44,071 tokens extracted from 4 documents.

The Masakhane project works mainly on low-resource MT for African languages, and is an excellent source of African language parallel corpora \cite{nekoto-etal-2020-participatory}. Currently, they have a growing online community of researchers who collaborate, contribute and share advancements in NLP for African languages. They have an open repository \footnote{\url{https://github.com/masakhane-io/masakhane-mt}} for sharing data as well as code and tools for building and facilitating NLP research in African languages. Nevertheless, the Emakhuwa language has not been explored yet by the community.

\begin{table*}[t]
\centering
\begin{tabular}{ L{2cm} L{4cm} L{4cm} L{4cm} }
\hline 
\textbf{Source} & \textbf{Portuguese} & \textbf{Emakhuwa} & \textbf{English} \\
\hline
 JW& depois, eu percebi que essa tinha sido a decisão certa. & nuuvira okathi kihoona wira vaari vooloka othaama. & After that, I realized that it was the right choice \\
 \hline
 Centro Catequ\'{e}tico Paulo VI & vivo aquilo em que acredito ? & vekekhai kinnettela seiyo sinà minaka ? &  Do I live on what I believe?\\
  \hline
 African Story Book& Certo homem tinha dois filhos & Mulopwana mmosa aahikhalana an'awe anli. & A man had two sons \\
 \hline
 Declaration of Human Rights & todos os seres humanos nascem livres e iguais em dignidade e em direitos . dotados de razão e de consciência , devem agir uns para com os outros em espírito de fraternidade . & atthu othene aniyaria oolikana ni owilamula moota ontthunaya okhala , variyari v edignidade ni edireito . akhalanne esaria ni otthokelela , ahaana akhalasaka othene saya vamurettele . &  the law needs to be known by the whole community, because from the knowledge of the law we can defend and safeguard our rights.\\
 \hline
 Land Law of Mozambique & nas áreas rurais , as comunidades locais participam : & mulaponi mwa imatta , ikomunidade sa elapo yeyo sokhalana mpantta : & in rural areas, local communities participate: \\
\hline
\end{tabular}
\caption{\label{examples} Examples from the Portuguese-Emakhuwa parallel corpus, with English translation. }
\end{table*}

\section{Data}
\label{data}

The dataset contains Portuguese-Emakhuwa parallel sentences, collected from crawling the Jehovah's Witness (JW) website\footnote{\url{https://www.jw.org}} and the African Story Book website\footnote{\url{https://www.africanstorybook.org}}. The African Story Book contains 17 short stories in Emakhuwa with Portuguese translations. The Jehovah's Witness website contains \textit{The Watchtower—Study Edition} magazine as well as \textit{The Watchtower} and \textit{Awaque!} from 2016 to 2021. Also, contains 27 books of the New Testament Bible.

The corpus also contains text from various institutional sources. These were digitized by performing optical character recognition (OCR) on a set of PDF files from Centro Catequ\'{e}tico Paulo VI (Paul VI Catechetical Centre), the Universal Declaration of Human Rights, and Mozambican Land Law. 
The majority of texts come from the Centro Catequ\'{e}tico Paulo VI which has published a series of three catechism books: \emph{Yesu Mopoli Ahu -- Jesus Nosso Salvador} (`Jesus Our Saviour'). These books, besides being religious, also touch upon a wide range of topics. 

The dataset contains texts written in the central Emakhuwa variant only, because this variant has relatively more resources than others. The sentences come from 125 pair documents in total, and amount to 699,976 word tokens of Emakhuwa.
Table~\ref{examples} shows examples from the dataset, with Portuguese and Emakhuwa parallel sentences and English glosses.

Processing of the OCR obtained from these documents involved the removal of some inconsistencies. For instance, the letter ``\`{i}'' was wrongly recognized as ``T'', and the letter ``\'{a}'' was recognized as ``6'', to give just two examples of OCR errors.  Processing also involved dividing longer sentences into shorter and concise sentences as well as discarding sentences wrongly recognized.
Then followed the alignment of Portuguese and Emakhuwa sentences which has been manually carried out.
More processing needs to be done, as it transpires that spelling standards vary between the data sources. As an example, there are some lexical conflicts in the corpus, particularly regarding the use of the letters ``j'' and ``c'' along with the suffix or infix ``erya''. JW texts tend to use the letter ``cerya'', while others use ``jerya''. So for example the English word ``begin'' would be written as ``opacerya'' in JW texts, whereas in others resources it would be written as ``opajerya''. Another example is the English word ``kiss'': its translation to Emakhuwa in the African Story Book is ``epeeco'', whereas in other resources is translated as ``epexo''.

\begin{table}[t]
\begin{tabular}{ L{1.2cm} L{1.2cm} R{2cm} R{1.5cm}}
\hline 
\textbf{Source} & \textbf{Lang} & \textbf{Sentences} & {\textbf{Tokens}} \\
\hline
JW & PT &  42,840 & 798,371 \\
  & VMW &   & 638,365\\
 \hline
CCA & PT &  4067 & 73,221\\
  & VMW &   & 56,724\\
  \hline
ASB & PT &  294 & 2746 \\
  & VMW &   & 1945\\
  \hline
LL & PT &  128 & 1656 \\
   & VMW &   & 1832 \\
  \hline
HR & PT &  86 & 1601 \\
  & VMW &   & 1110\\
     \hline
 \textbf{Total}& PT & 47,415 & 877,595 \\
 & VMW & & 699,976 \\
\hline
\end{tabular}
\caption{\label{counts} Dataset counts where: JW = Jehovah's Witness, CCA = Centro Catequ\'{e}tico Paulo VI, ASB = African Book Story, LL = Land Law of Mozambique, HR = Universal Declaration of Human Rights; PT = Portuguese, VMW = Emakhuwa. }
\end{table}

Word borrowing is another issue to take into account, since Emakhuwa takes many words from Portuguese, but without a consensus on spelling conventions. As an example, take the word ``bible'' which in Portuguese is "bíblia". In Emakhuwa it is sometimes written as ``biblia'' and at other times written as ``biibiliya''.
This inconsistency also exists within Emakhuwa language resources itself, as the alphabet and spelling standards went through several revisions -- the latest in 2012 \cite{ngunga-faquir-2012}.  
Therefore, some future processing is necessary to normalize the word standards in the dataset. The corpus will be available for research use, and updates may be obtained from our project website\footnote{\url{placeholder.com}}.
Table~\ref{counts} summarizes the corpus in terms of sentences and the number of tokens coming from each data source.

\section{Baseline translation model}

We train an initial NMT model with the default OpenNMT configuration\footnote{https://github.com/OpenNMT/OpenNMT-py/blob/master/docs/source/quickstart.md}, which consists of a 2-layer LSTM with 500 hidden units on both the encoder and decoder \cite{opennmt}. Sentences from the corpus are randomly assigned to training and test sets at a ratio of 9:1. These data splits will be released with the corpus so that others may compare the performance of their models against this one.
We evaluate with BLEU \cite{bleu} as shown in Table ~\ref{results}. This presents a baseline level of performance to improve upon in future work. 

\begin{table}[t]
\begin{tabular}{ L{2cm} R{2cm} R{2cm}}
\hline
 \multicolumn{2}{c}{}{\textbf{Target}} \\
\hline 
\textbf{Source} & \textbf{Portuguese}& {\textbf{Emakhuwa}} \\
\hline
\textbf{Portuguese} & -  & 12.42\\
\textbf{Emakhuwa} & 13.09 & -\\
\hline
\end{tabular}
\caption{\label{results} BLEU scores for Emakhuwa-Portuguese and Portuguese-Emakhuwa NMT models. }
\end{table}

\section{Conclusion}

In this paper we describe the creation of a new parallel corpus of Emakhuwa and Portuguese. The preparation of the corpus is on-going as the texts require more processing and normalization, but it will be made freely available for research use, most probably for machine translation. Currently, the dataset is made up of mostly religious and legal content but in future our objective is to diversify the range of sources and topics covered. This is important in order that Portuguese-Emakhuwa translation models work well across various domains.

% do not include for review
\section*{Acknowledgements}
The authors wish to acknowledge the Centro Catequ\'{e}tico Paulo VI for allowing to use their materials for research purpose.  
The second author is supported by Research England via the University of Cambridge Global Challenges Research Fund. He wishes to thank Paula Buttery, Tanya Hall, Carol Nightingale \& Sara Serradas Duarte for their help and guidance.

\bibliography{eacl2021}
\bibliographystyle{acl_natbib}

\end{document}